\documentclass{article}

\usepackage[top=1in, bottom=1in, left=1in, right=1in]{geometry}
\usepackage{paper}

\newcommand{\para}[3]{{#1}^{#2}\big(#3\big)}
\newcommand{\X}[1]{\para{X}{}{#1}}
\newcommand{\Y}[1]{\para{Y}{}{#1}}
\newcommand{\Z}[3]{\para{Z}{(#1)}{#2, #3}}
\newcommand{\W}[4]{\para{W}{(#1)}{#2, #3, #4}}
\newcommand{\D}[5]{\para{D}{(#1)}{#2, #3, #4, #5}}
\newcommand{\B}[2]{\para{B}{(#1)}{#2}}

\newcommand{\exeq}{\overset{\diamond}{=}}

\newcommand{\ie}{i.e.}
\newcommand{\BNetT}{BNet2}

\title{Butterfly-Net2: Simplified Butterfly-Net and Fourier
Transform Initialization}

\author{
    Zhongshu Xu$^{\,\dagger}$,
    Yingzhou Li$^{\,\sharp\,\star}$,
    Xiuyuan Cheng$^{\,\sharp}$
    \vspace{0.1in}\\
    $\dagger$ School of Mathematical Sciences, University of Science
    and Technology of China\\
    $\sharp$ Department of Mathematics, Duke University\\
    $\star$ yingzhou.li@duke.edu
}

\begin{document}

\maketitle

\begin{abstract}
Structured CNN designed using the prior information of problems
potentially improves efficiency over conventional CNNs in various
tasks in solving PDEs and inverse problems in signal processing. This
paper introduces BNet2, a simplified Butterfly-Net and inline with
the conventional CNN.  Moreover, a Fourier transform initialization
is proposed for both BNet2 and CNN with guaranteed approximation
power to represent the Fourier transform operator. Experimentally,
BNet2 and the Fourier transform initialization strategy are tested on
various tasks, including approximating Fourier transform operator,
end-to-end solvers of linear and nonlinear PDEs, and denoising and
deblurring of 1D signals. On all tasks, under the same initialization,
BNet2 achieves similar accuracy as CNN but has fewer parameters. And
Fourier transform initialized BNet2 and CNN consistently improve the
training and testing accuracy over the randomly initialized CNN.
\end{abstract}


\section{Introduction}

Deep convolutional neural network (CNN) has been widely applied
to solving PDEs as well as inverse problems in signal processing.
In both applications, spectral methods, namely involving forward and
backward Fourier transform operators, serve as a traditional solution.
Spectral methods have been a classical tool for solving elliptic PDEs.
For image inverse problems, primarily image restoration like denoising
and deblurring, a large class of PDE methods consider the nonlinear
diffusion process~\cite{perona1990scale,tsiotsios2013choice} which are
connected to wavelet frame methods~\cite{cai2012image,dong2017image}.
The involved operator is elliptic, typically the Laplace operator.

Apart from the rich prior information in these problems, the
conventional end-to-end deep CNN, like U-Net~\cite{ronneberger2015u}
and Pix2pix~\cite{isola2017image}, consists of convolutional layers
which have fully trainable local filters and are densely connected
across channels. The enlarged model capacity and flexibility of
CNN improves the performance in many end-to-end tasks, however,
such fully data-driven models may give a superior performance
on one set of training and testing datasets, but encounter
difficulty when transfer to another dataset, essentially due to
the overfitting of the trained model which has a large amount of
flexibility~\cite{plotz2017benchmarking,abdelhamed2018high}.  Also,
as indicated in~\cite{Chollet2017, Jin2015, Mamalet2012, Wang2017,
Wang2018c}, the dense channel connection can be much pruned in the
post-training process without loss of the prediction accuracy.

This motivates the design of structured CNNs which balance between
model capacity and preventing over-fitting, by incorporating prior
knowledge of PDEs and signal inverse problems into the deep CNN
models.  The superiority of structured CNNs over ordinary CNNs has
been shown both for PDE solvers~\cite{fan2019multiscale-hmat}
and for image inverse problems~\cite{gilton2019neumann}.
Several works have borrowed ideas from numerical analysis in deep
models: \cite{Li2019} introduces Butterfly-Net (BNet) based on
the butterfly algorithm for fast computation of Fourier integral
operators~\cite{Ying2009,Candes2009,Demanet2012a,Li2015a,Li2015b};
\citet{khoo2019switchnet} proposes to use a switching layer with sparse
connections in a shallow neural network, also inspired by the butterfly
algorithm, to solve wave equation based inverse scattering problems;
\citet{fan2019multiscale-hmat} and \citet{fan2019multiscale-nested}
introduce hierarchical matrix into deep network to compute nonlinear
PDE solutions; \citet{fan2019bcr} proposes a neural network based on
the nonstandard form~\cite{beylkin1991fast} and applies to approximate
nonlinear maps in PDE computation.  Problem-prior informed structured
CNNs have become an emerging tool for a large class of end-to-end
tasks in solving PDEs and inverse problems.

This paper introduces a new Butterfly network architecture,
which we call \BNetT{}. A main observation is that, as long as
the Fourier transform operator is concerned, the switch layer in
Butterfly algorithm can be removed while preserving the approximation
ability of BNet. This leads to the proposed model, which inherits
the approximation guarantee the same as the BNet, but also makes the
network architecture much simplified and inline with the conventional
CNN.

We also investigate the Fourier transform (FT) initialization. FT
initialization adopts the interpolative construction in Butterfly
factorization as in~\cite{Li2017} to initialize \BNetT{}.  Since
\BNetT{} now is a conventional CNN with sparsified channel connections,
FT initialization can also be applied to CNN to realize a linear
FT. We experimentally find that both \BNetT{} and CNN are sensitive
to initialization in problems that we test on, and FT initialized
networks outperform their randomly initialized counterpart in our
settings. The trained network from FT initialization also demonstrates
better stability with respect to statistical transfer of testing
dataset from the training set.

In summary, the contributions of the paper include: (1) We introduce
\BNetT{}, a simplified structured CNN based on Butterfly algorithm,
which removes the switch layer and later layers in BNet and thus is
inline with the conventional CNN architecture; (2) FT initialization
for both \BNetT{} and CNN serves as an initialization recipe for a
large class of CNNs in many applications; (3) FT initialized \BNetT{}
and CNN inherit the theoretical exponential convergence of BNet in
approximating FT operator, and the approximation can be further
improved after training on data; (4) Applications to end-to-end
solver for linear and nonlinear PDEs, and inverse problems of signal
processing are numerically tested, and under the same initialization
\BNetT{} with fewer parameters achieves similar accuracy as CNN;
(5) FT initialized \BNetT{} and CNN outperforms randomly initialized
CNN on all tasks included in this paper.

\section{Butterfly-Net2}

The structure of \BNetT{} inherits a part of design in BNet but makes
it more simple and similar to CNN.  The specific difference between
BNet and \BNetT{} will be presented in Remark~\ref{rmk:BNetBNet2}.  For
completeness, we will first recall the CNN under our own notations and
then introduce \BNetT{}. Towards the end of this section, the numbers
of parameters for both CNN and \BNetT{} are derived and compared.

Before introducing network structures, we first familiarize ourselves
with notations used throughout this paper. The bracket notation of
$n$ denotes the set of nonnegative integers smaller than $n$, \ie,
$[n] = \{0, 1, \dots, n-1\}$. Further, a contiguous subset of $[n]$
is denoted as $[n]^k_i$, where $k$ is a divisor of $n$ denoting the
total number of equal partitions and $i$ indexed from zero denotes the
$i$-th partition, \ie, $[n]^k_i = \{\frac{n}{k}i, \frac{n}{k}i + 1,
\dots, \frac{n}{k}(i+1) - 1\}$. While describing the network structure,
$X$ and $Y$ denote the input and output vector with length $N$ and
$K$ respectively, \ie, $X = \X{[N]}$ and $Y = \Y{[K]}$. $Z$, $W$,
and $B$ denote hidden variables, multiplicative weights and biases
respectively. For example, $\Z{\ell}{[n]}{[C]}$ is the hidden variable
at $\ell$-th layer with $n$ spacial degrees of freedom (DOFs) and $C$
channels; $\W{\ell}{[w]}{[C_{in}]}{[C_{out}]}$ is the multiplicative
weights at $\ell$-th layer with $w$ being the kernel size, $C_{in}$ and
$C_{out}$ being the in- and out-channel sizes; $\B{\ell}{[C_{out}]}$
denotes the bias at $\ell$-th layer acting on $C_{out}$ channels.
Activation function is denoted as $\sigma(\cdot)$, which is ReLU in
this paper by default.

\subsection{CNN Revisit}

A one dimensional CNN can be precisely described using notations
defined as above. For a $L$ layer CNN, we define the feedforward
network as follows.
\begin{itemize}
    \item \textbf{Layer 0: } The first layer hidden variable
    $\Z{1}{[\frac{N}{2w}]}{[2r]}$ with $2r$ channels is generated
    via applying a 1D convolutional layer with kernel size $2w$
    and stride $2w$ followed by an activation function on the input
    vector $\X{[N]}$, \ie,
    \begin{equation} \label{eq:CNNlayer0}
        \Z{1}{j}{c} = \sigma \Big(\B{0}{c} + \sum_{i \in [2w]}
        \W{0}{i}{0}{c} \X{2wj+i} \Big),
    \end{equation}
    for $j \in [\frac{N}{2w}]$ and $c \in [2r]$.

    \item \textbf{Layer $\ell = 1, 2, \dots, L-1$: } The connection
    between the $\ell$-th layer and the $(\ell+1)$-th layer hidden
    variables is a 1D convolutional layer with kernel size $2$,
    stride size $2$, $2^\ell r$ in-channels and $2^{\ell+1} r$
    out-channels followed by an activation function, \ie,
    \begin{equation} \label{eq:CNNlayer}
        \Z{\ell+1}{j}{c} = \sigma \Big(\B{\ell}{c} + \sum_{k
        \in [2^\ell r]} \sum_{i \in [2]} \W{\ell}{i}{k}{c}
        \Z{\ell}{2j+i}{k} \Big),
    \end{equation}
    for $j \in [\frac{N}{2^{\ell+1} w}]$ and $c \in [2^{\ell+1} r]$.
    The first and second summation in \eqref{eq:CNNlayer} denotes
    the summation over in-channels and the spacial convolution
    respectively.

    \item \textbf{Layer $L$: } The last layer mainly serves as
    a reshaping from the channel direction to spacial direction,
    which links the $L$-th layer hidden variables with the output $Y$
    via a fully connected layer, \ie,
    \begin{equation} \label{eq:CNNlayerL}
        \Y{c} = \sum_{k \in [2^L r]} \sum_{i \in [\frac{N}{2^L w}]}
        \W{L}{i}{k}{c} \Z{L}{i}{k},
    \end{equation}
	for $c \in [K]$. If $Y$ is not the final output, then the
	bias and activation function can be added.

\end{itemize}
In the above description, readers who are familiar with CNN may find
a few irregular places. We will address these irregular places in
Remark~\ref{rmk:CNNBNet} after the introduction of \BNetT{}.

CNN is the most successful neural network in practice, especially
in the area of signal processing and image processing. Convolutional
structure, without doubt, contributes most to this success. Another
contributor is the increasing channel numbers. In practice, people
usually double the channel numbers until reaching a fixed number
and then stick to it till the end.  Continually doubling channel
numbers usually improves the performance of the CNN, but has two
drawbacks. First, large channel numbers lead to the large parameter
number, which in turn leads to overfitting issue. The second drawback
is the expensive computational cost in both training and evaluation.

\begin{figure}[t]
    \centering \subfigure[Network Structure]{
    \includegraphics[width=0.6\textwidth]{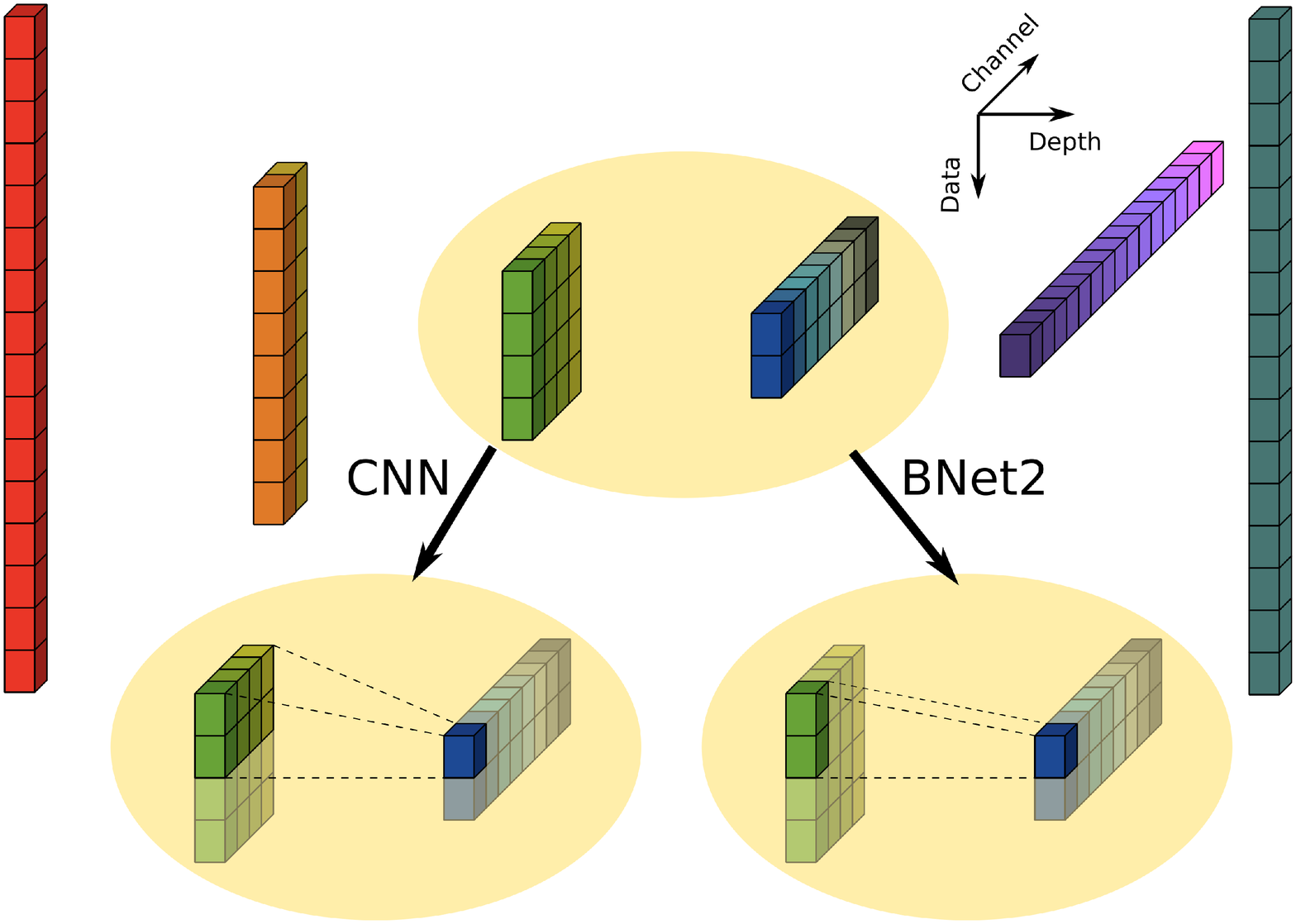}}
    ~~~~ \subfigure[Channel Connectivity]{
    \includegraphics[width=0.33\textwidth]{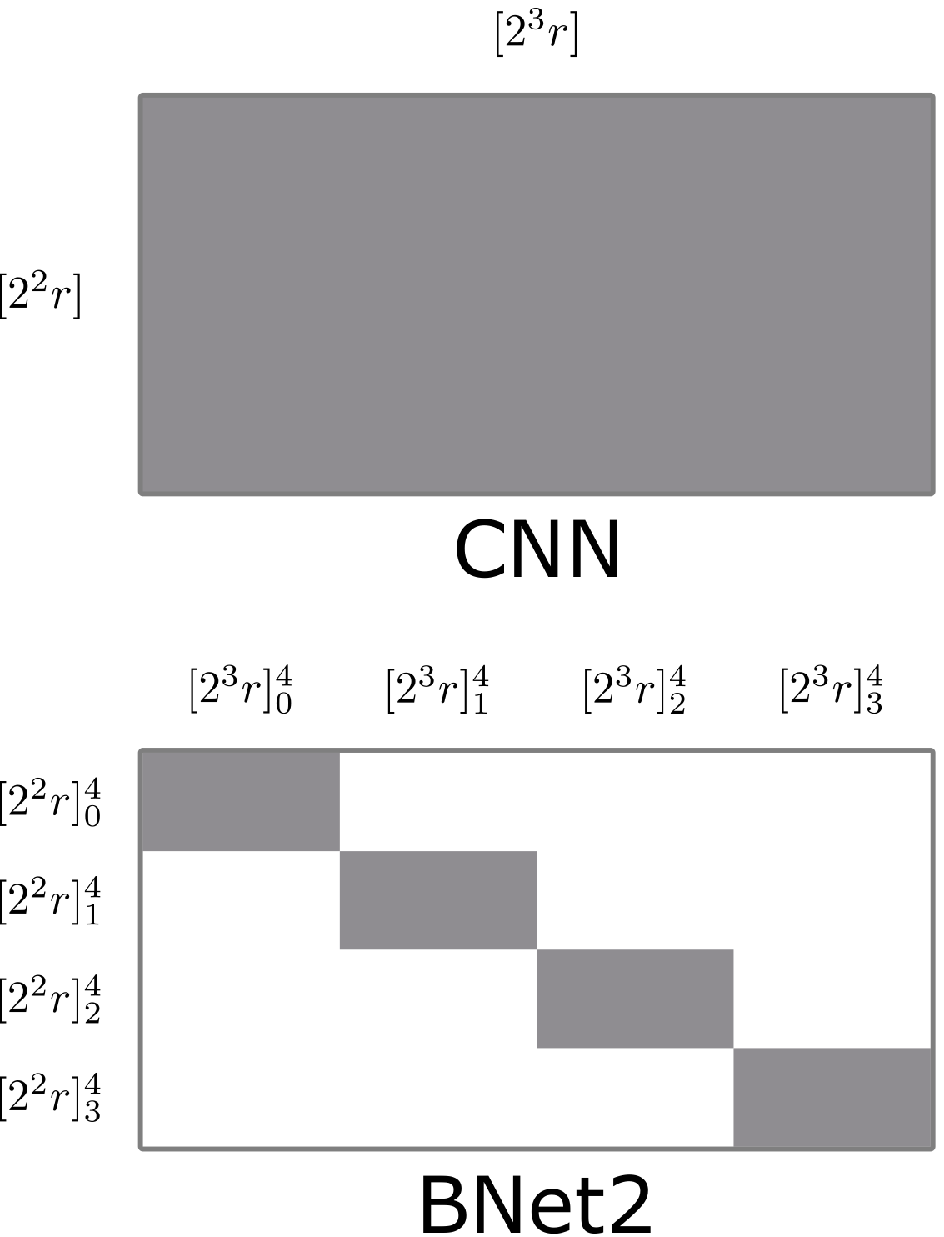}}
    \caption{The comparison between CNN and \BNetT{}.}
    \label{fig:CNNBNet}
\end{figure}

\subsection{Butterfly-Net2}
\label{sec: BNet2}

\BNetT{}, in contrast to CNN, has the identical convolutional structure
and allows continually doubling the channel numbers.  For the two
drawbacks mentioned above, \BNetT{} overcomes the second one and
partially overcomes the first one.

The Layer 0 in \BNetT{} is identical to that in CNN. Hence, we only
define other layers in the feed forward network as follows.
\begin{itemize}
    \item \textbf{Layer $\ell = 1, 2, \dots, L-1$: } The $2^\ell r$
    in-channels are equally partitioned into $2^\ell$ parts. For
    each part, a 1D convolutional layer with kernel size $2$, stride
    $2$, in-channel size $r$ and out-channel size $2r$ is applied.
    The connection between the $\ell$-th layer and the $(\ell+1)$-th
    layer hidden variables obeys,
    \begin{equation} \label{eq:BNetlayer}
        \Z{\ell+1}{j}{c} = \sigma \Big(\B{\ell}{c} + \sum_{k \in
        [2^\ell r]^{2^\ell}_p} \sum_{i \in [2]} \W{\ell}{i}{k}{c}
        \Z{\ell}{2j+i}{k} \Big),
    \end{equation}
    for $j \in [\frac{N}{2^{\ell+1} w}]$, $c \in [2^{\ell+1}
	r]^{2^\ell}_p$, and $p \in [2^\ell]$.

    \item \textbf{Layer $L$: } The $2^Lr$ in-channels
	are equally partitioned into $2^L$ parts. For each part,
	a 1D convolutional layer with kernel size $\frac{N}{2^Lw}$,
	in-channel size $r$ and out-channel size $\frac{K}{2^L}$
	is applied. The last layer links the $L$-th layer hidden
	variables with the output $Y$,
    \ie,
    \begin{equation} \label{eq:BnetlayerL}
        \Y{c} = \sum_{k \in [2^L r]^{2^L}_p} \sum_{i \in [\frac{N}{2^L
        w}]} \W{L}{i}{k}{c} \Z{L}{i}{k},
    \end{equation}
	for $c \in [K]^{2^L}_p$ and $p \in [2^L]$. If $Y$ is not
	used as the final output directly, then the bias term and
	activation function
    can be added.
\end{itemize}

\begin{remark} \label{rmk:CNNBNet}
    This remark addresses two irregular places in the CNN and
    \BNetT{} described above against the conventional CNN.  First,
    all convolutions are performed in a non-overlapping way, \ie,
    the kernel size equals the stride size. Regular convolutional
    layer with a pooling layer can be adopted to replace the
    non-overlapping convolutional layer in both CNN and \BNetT{}.
    Second, except the Layer 0 and Layer $L$, all kernel sizes are
    $2$, which can be generalized to other constant for both CNN and
    \BNetT{}.  We adopt such presentations to simplify the notations
    in Section~\ref{sec:FTinit}, the FT initialization.
\end{remark}

\begin{remark} \label{rmk:BNetBNet2}
    This remark addresses the difference between BNet and \BNetT{}. As
    seen in Section~\ref{sec: BNet2} and Appendix~\ref{sec:BNet},
    \BNetT{} deletes Switch Layer and Conv-T Layers in BNet and expands
    Conv Layers to all layers. So the Switch Layer and Layer L in
    BNet are actually combined to be the Layer L in \BNetT{}. These
    difference make the structure of \BNetT{} simpler than BNet. On
    the other hand, the removal of Switch Layer and Conv-T Layers
    makes \BNetT{} directly a sparsified regular CNN, while BNet does
    not has such a property.
\end{remark}

In \eqref{eq:BNetlayer}, the in-channel index $k$ and the out-channel
index $c$ of \BNetT{} are linked through the auxiliary index $p$,
whereas in the CNN, the in-channel index $k$ and the out-channel index
$c$ are independent (see \eqref{eq:CNNlayer}). Figure~\ref{fig:CNNBNet}
(b) illustrates the connectivity of $k$ and $c$ in CNN and in
\BNetT{} at the $2$-nd layer. Further, Figure~\ref{fig:CNNBNet} (a)
shows the overall structure of CNN and \BNetT{}. If we fill part of
the multiplicative weights in CNN to be that in \BNetT{} according
to \eqref{eq:BNetlayer} and set the rest multiplicative weights to
be zero, then CNN recovers \BNetT{}.  Hence, any \BNetT{} can be
represented by a CNN. The approximation power of CNN is guaranteed
to exceed that of \BNetT{}. Surprisingly, according to our numerical
experiments, the extra approximation power does not improve the
training and testing accuracy much in all examples we have tested.

\subsection{Parameter Counts}

Parameter counts are explicit for both CNN and \BNetT{}. The numbers
of bias are identical for two networks. It is $2r$ for Layer 0,
$2^{\ell+1} r$ for Layer $\ell$ and $0$ for Layer $L$.  Hence the
overall number of biases is
\begin{equation}
    \sharp \{\text{bias}\} = \sum_{\ell \in [L]} 2^{\ell+1} r =
    (2^{L+2}-2)r.
\end{equation}
The total number of multiplicative parameters are very different
for CNN and \BNetT{}.  For CNN, the parameter count is $2r \cdot 2w$
for Layer 0, $2^\ell r \cdot 2^{\ell+1} r \cdot 2$ for Layer $\ell$,
and $2^L r \cdot K \cdot \frac{N}{2^L w}$ for Layer $L$.  Hence the
overall number of multiplicative parameters for CNN is
\begin{equation}
    \sharp \{W_{CNN}\} = 4rw + \frac{rNK}{2^Lw} + \sum_{\ell = 1}^{L-1}
    2^{2\ell+2} r^2 = 4rw + \frac{rNK}{w} + \frac{2^{2L+2}-2^4}{3}r^2.
\end{equation}
While, for \BNetT{}, the parameter count is $2r \cdot 2w$ for Layer
0, $2^\ell r \cdot 2r \cdot 2$ for Layer $\ell$, and $2^L r \cdot
\frac{K}{2^L} \cdot \frac{N}{2^L w}$ for Layer $L$.  Hence the overall
number of multiplicative parameters for \BNetT{} is
\begin{equation}
    \sharp \{W_{\BNetT{}}\} = 4rw + \frac{rNK}{2^Lw} + \sum_{\ell
    = 1}^{L-1} 2^{\ell+2} r^2 = 4rw + \frac{rNK}{2^Lw} +
    \frac{2^{L+2}-2^3}{3}r^2.
\end{equation}
If we assume $N \sim K \sim 2^L$ and $r \sim w \sim 1$, which
corresponds to doubling channel number till the end, then we have
\begin{equation}
    \sharp\{\text{parameter in CNN}\} = O(N^2) \quad \text{and}
    \quad \sharp\{\text{parameter in \BNetT{}}\} = O(N).
\end{equation}
Let us consider another regime, i.e., $K \sim 2^L$ and $r \sim
\frac{N}{2^Lw} \sim 1$, which can be viewed as an analog of doubling
the channel number first and then being fixed to a constant $2^L$.
Under this regime, the total numbers of parameters can be compared as,
\begin{equation}
    \sharp\{\text{parameter in CNN}\} = O(\frac{N}{K} + K^2) \quad
    \text{and} \quad \sharp\{\text{parameter in \BNetT{}}\} =
    O(\frac{N}{K} + K),
\end{equation}
where both $\frac{N}{K}$ come from $4rw$ term.  Hence, in both
regimes of hyper parameter settings, \BNetT{} has lower order number
of parameters comparing against CNN. If the performance in terms of
training and testing accuracy remains similar, \BNetT{} is then much
more preferred than the CNN.

\section{Fourier Transform Initialization}
\label{sec:FTinit}

A good initialization is crucial in training CNNs especially in
training highly structured neural networks like \BNetT{}. It
is known that CNN with random initialization achieves
remarkable results in practical image processing tasks as shown
in~\cite{Krizhevsky2012}. However, for synthetic signal data as in
Section~\ref{sec:numerical}, in which the high accuracy prediction
is possible through a CNN with a set of parameters, we notice that
CNN with random initialization and ADAM stochastic gradient descent
optimizer is not able to converge to that CNN.

In this section, we aim to initialize both \BNetT{} and CNN to fulfill
the discrete FT operator, which is defined as
\begin{equation} \label{eq:DFT}
    \calK(\xi,t) = \Kfun{\xi \cdot t},
\end{equation}
for $\xi \in [K]$ and $t \in \frac{[N]}{N}$, where $N$ denotes
the number of discretization points and $K$ denotes the frequency
window size. When the network is initialized as an approximated
discrete FT, we call it \textbf{FT initialization}. Discrete FT is
the traditional computational tool for signal processing and image
processing. Almost all related traditional algorithms involve either
FT directly or Laplace operator, which can be realized via two FTs,
see \cite{Buades2005, Chan2005}.  Hence, if we can initialize a
neural network as such a traditional algorithm involving discrete FT,
the training of the neural network would be viewed as refining the
traditional algorithm and makes its data adaptive.  In another word,
neural network solving image processing and signal processing tasks
can be guaranteed to outperform traditional algorithms, although it is
widely accepted in practice.  This section is composed of two parts:
preliminary and initialization.  We will first introduce related
complex number neural network realization, Chebyshev interpolation,
FT approximation in the preliminary part.  The receipt of the
FT initialization for both \BNetT{} and CNN is then introduced
in detail, mainly in \eqref{eq:FTinit1}, \eqref{eq:FTinit2}, and
\eqref{eq:FTinit3}.

\subsection{Preliminary}

Fourier transform is a linear operator with complex coefficients.
The realization of complex number operations via ReLU neural network
is detailed in Appendix~\ref{sec:complexnetwork} together with the
definition of the extensive assign operator $\exeq$.

An important tool is the Lagrange polynomial on Chebyshev points.
The Chebyshev points of order $r$ on $[-\frac{1}{2}, \frac{1}{2}]$
is defined as,
\begin{equation}
    \bigg\{z_i = \frac{1}{2} \cos \Big( \frac{i\pi}{r}
    \Big) \bigg\}_{i\in [r]}.
\end{equation}
The associated Lagrange polynomial at $z_k$ is
\begin{equation}
    \calL_k(x) = \prod_{p \neq k}\frac{x-z_p}{z_k-z_p}.
\end{equation}
If the interval $[-\frac{1}{2}, \frac{1}{2}]$ is re-centered at $c$
and scaled by $w$, then the transformed Chebyshev points obeys $z'_i =
wz_i+c$ and the corresponding Lagrange polynomial at $z'_k$ is
\begin{equation}
    \calL'_k(x) = \prod_{p \neq k}\frac{x-z'_p}{z'_k-z'_p}
    = \prod_{p \neq k}\frac{\frac{x-z'_k}{w} + z_k-z_p}{z_k-z_p}
    =\widetilde{\calL}_k(\frac{x-z'_k}{w}),
\end{equation}
where $\widetilde{\calL}_k(\cdot)$ is independent of the transformation
of the interval.

Recall the Chebyshev interpolation representation of FT as
Theorem~2.1 in~\cite{Li2019}. We include part of that theorem with
a small modification here with our notation $\widetilde{\calL}_k$
for completeness.

\begin{proposition}[Theorem 2.1 in~\cite{Li2019}] \label{thm:low-rank}
    Let $L$ and $r$ be two parameters such that $\pi e K \leq r 2^{L}$.
    For any $\ell \in [L]$, let $A^{\ell+1}$ and $B^{L-\ell-1}$ denote
    two connected subdomains of $[-\nicefrac{K}{2}, \nicefrac{K}{2})$
    and $[0,1)$ with length $K\cdot 2^{-\ell-1}$ and $2^{\ell+1-L}$
    respectively.  Then for any $\xi \in A^{\ell+1}$ and $t
    \in B^{L-\ell-1}$, there exists a Chebyshev interpolation
    representation of the FT operator,
    \begin{equation} \label{eq:low-rank1}
        \abs{\,\Kfun{\xi \cdot t} - \sum_{k = 1}^{r}\Kfun{\xi
        \cdot t_k} \Kfun{\xi_0 \cdot \left( t - t_k \right)}
        \widetilde{\calL}_k\Big(\frac{t-t_k}{2^{L-\ell-1}}\Big)\,}
        \leq \left( 2 + \frac{2}{\pi} \ln r \right)
        \left( \frac{ \pi e K }{ r 2^{L+1}} \right)^{r},
    \end{equation}
    where $\xi_0$ is the centers of $A^{\ell+1}$, $\{t_k\}_{k \in [r]}$
    are the Chebyshev points on $B^{L-\ell-1}$.
\end{proposition}
Obviously, part of the approximation, $\Kfun{\xi_0 \cdot \left( t -
t_k \right)} \widetilde{\calL}_k\Big(\frac{t-t_k}{2^{L-\ell-1}}\Big)$,
admits the convolutional structure across all $B^{L-\ell-1}$. This part
will be called the \textbf{interpolation part} in the following. It
is the key that we can initialize CNN and \BNetT{} as a FT.

\subsection{Fourier Transform Initialization for CNN and BNet2}

Since all weights fit perfectly into the structure of \BNetT{}, we will
only introduce the initialization of \BNetT{} in detail. Assume the
input is a function discretized on a uniform grid of $[0,1)$ with $N$
points and the output is the discrete FT of the input at frequency
$[K]$. Throughout all layers, the bias terms are initialized with
zero. In the description below, we focus on the initialization of
the multiplicative weights. Without loss of generality, we further
assume $N = w2^L$.
\begin{itemize}
    \item \textbf{Layer 0: } For $\ell = 0$, we consider $A^1_i =
    [K]^2_i$ and $B^{L-1}_j = \frac{[N]^{2^{L-1}}_j}{N}$ for $i \in
    [2]$ and $j \in [2^{L-1}]$, which satisfies the condition in
    Proposition~\ref{thm:low-rank}. An index pair $(i,k)$ for $i$
    being the index of $A^1_i$ and $k$ being the index of the Chebyshev
    points can be reindexed as $[2r]$. Hence we abuse the channel index
    $c$ as $c=(i,k)$. Then fixing $c$, the interpolation part is the
    same for all $B^{L-1}_j$, which is naturally a non-overlapping
    convolution. Hence we set
    \begin{equation} \label{eq:FTinit1}
        \W{0}{Nt}{0}{c} \exeq \Kfun{\xi_0 \cdot
	    \left( t - t_k \right)}
	    \widetilde{\calL}_k\Big(\frac{t-t_k}{2^{L-1}}\Big)
    \end{equation}
    for $t \in \frac{[\frac{N}{2^{L-1}}]}{N}$, $c = (i,k)$, and $t_k$
    is the Chebyshev point on $B^{L}_0$ or $B^{L}_1$.  Then after
    applying the 1D convolutional layer as \eqref{eq:CNNlayer0},
    the first hidden variable $\Z{1}{j}{c}$ represents the input
    vector interpolated to the Chebyshev points $t_k$ on $B^{L-1}_j$
    with respect to $A^1_i$. The following layers recursively apply
    Proposition~\ref{thm:low-rank} to the remaining $\Kfun{\xi\cdot
    t_k}$ part.

    \item \textbf{Layer $\ell = 1, 2, \dots, L-1$: } We concern
    $A^{\ell+1}_i$ and $B^{L-\ell-1}_j$ for $i \in [2^{\ell+1}]$
    and $j \in [2^{L-\ell-1}]$ at the current layer. The hidden
    variable $\Z{\ell}{j'}{c'}$ represents the input interpolated
    to the Chebyshev points on $B^{L-\ell}_{j'}$ with respect to
    $A^{\ell}_{i'}$, where $c' = (i',k')$ and $k'$ is the index
    of Chebyshev points. $A^{\ell+1}_{i}$ is a subinterval of
    $A^{\ell}_{\floor{\frac{i}{2}}}$ and $B^{L-\ell-1}_j$ covers
    $B^{L-\ell}_{2j}$ and $B^{L-\ell}_{2j+1}$.  For a fixed $c=(i,k)$,
    the interpolation part is the same for each index $j$. The
    convolution kernel, hence, is defined as,
    \begin{equation} \label{eq:FTinit2}
        \W{\ell}{p}{c'}{c} \exeq \Kfun{\xi_0
        \cdot \left( t_{k'} - t_k \right)}
        \widetilde{\calL}_k\Big(\frac{t_{k'}-t_k}{2^{L-\ell-1}}\Big)
    \end{equation}
    where $c' = (\floor{\frac{i}{2}},k')$, $c = (i,k)$, $t_{k'}$
    and $t_k$ are Chebyshev points on $B^{L-\ell}_{2j+p}$ and
    $B^{L-\ell-1}_{j}$ respectively, and $p \in [2]$.
        
    \item \textbf{Layer $L$: } This layer concerns $A^L_i$ and
    $B^0_0$ for $i \in [2^L]$. All previous layers take care of the
    interpolation part. And the current layer applies the FT operator
    on each $A^L_i$.  The hidden variable $\Z{L}{j}{c'}$ represents
    the input interpolated to the Chebyshev points on $B^{L}_{0}$
    with respect to $A^{\ell}_{i}$, where $c' = (i,k')$ and $k'$ is
    the index of Chebyshev points.  Define the channel index $c$ as
    an index pair $(i,k)$, where $i \in [2^L]$ is the index of $A^L_i$
    and $k \in [\frac{K}{2^L}]$ is the index for uniform points $\xi_k
    \in A^L_i$.  Then the multiplicative weights are initialized as,
    \begin{equation} \label{eq:FTinit3}
        \W{L}{0}{c'}{c} \exeq \Kfun{\xi_k \cdot t_{k'}}.
    \end{equation}
\end{itemize}
Since \BNetT{} can be viewed as a CNN with many zero weights, such
an initialization can be used to initialize CNN as well. When we set
the weights as above and set the rest weights to be zero, the CNN is
then initialized by the FT initialization.

\begin{remark} \label{rmk:FT}
    As mentioned in Remark~\ref{rmk:CNNBNet}, a few irregular
    places in the current CNN and \BNetT{} description can be
    modified to match conventional CNN.  The FT initialization can
    be updated accordingly.  First, when convolutions are performed
    in a non-overlapping way without pooling layer, we can enlarge
    the kernel size and embed zeros to eliminate the impact of the
    overlapping part.  Second, when the kernel sizes are a constant
    different from $2$, the generalization of the initialization
    is feasible as long as the bipartition is modified to a
    multi-partition.
\end{remark}

The approximation power of FT initialized CNN and \BNetT{} can be
analyzed in an analogy way as that in~\cite{Li2019} and the sketch
proof in Appendix~\ref{sec:proof-matrix-approx}.

\begin{theorem} \label{thm:matrix-approx}
    Let $N$ and $K$ denote the size of the input and output
    respectively.  The depth $L$ and channel parameter $r$ satisfies
    $\pi e K \leq r 2^{L}$. Then there exists a \BNetT{}/CNN,
    $\calB(\cdot)$, approximating the discrete FT operator such that
    for any bounded input vector $f$, the error satisfies,
    \begin{equation}\label{eq:bound-thm5}
        \norm{ \calK f - \calB(f) }_p \leq C_{r,K} \left(
        \frac{r^{1-\frac{1}{p}} \left( \frac{2}{\pi}\ln r + 1
        \right)}{2^{r-2}} \right)^L \norm{f}_p,
    \end{equation}
    where $C_{r,K} = \nicefrac{(2+\nicefrac{4}{\pi}\ln r)^3(\pi
    eK)^r}{(2r)^{r-1}}$ is a constant depending only on $r$ and $K$,
    for $p \in [1,\infty]$.
\end{theorem}

Theorem~\ref{thm:matrix-approx} is validated numerically in next
section.  In terms of function approximation, \citet{Li2019} showed
that for BNet a result in the type of \eqref{eq:bound-thm5} implies
that a large class of functions can be well approximated with network
complexity depending on the effective frequency bandwidth instead of
the input dimension.  Based on Theorem~\ref{thm:matrix-approx},
such a function approximation result applies to \BNetT{} and
CNN as well. The approximation analysis can be extended to
vector-valued output functions. Numerically, we apply \BNetT{}
to the computation of the Laplace operator energy, which has
scalar output (Section \ref{sec:energyLaplace}) , as well as
end-to-end solvers of PDEs (Section \ref{sec:nonlinearequation},
\ref{sec:nonlinearequation}) and signal processing inverse problems
(Section \ref{subsec:exp-denoise-deblur}), both of which have
vector-valued output.

\section{Numerical Results}
\label{sec:numerical}

This section presents numerical experiments to demonstrate the
approximation power of CNN and \BNetT{}, and compare the difference
between FT initialization and random initialization.  Thus, four
different settings, CNN with random initialization (CNN-rand),
CNN with FT initialization (CNN-FT)\footnote{The Layer $L$ is
often combined with feature layers. Hence for both CNN, Layer $L$
as in \BNetT{} is adopted.}, \BNetT{} with random initialization
(\BNetT{}-rand), and \BNetT{} with FT initialization (\BNetT{}-FT)
are tested on three different sets of problems: (1) approximation
of FT operator; (2) energy and solution maps of elliptic equations;
(3) 1D signals de-blurring and de-noising tasks.

\subsection{Approximation of Fourier Transform Operator}

This section repeats experiments as in the original BNet~\cite{Li2019}
on \BNetT{}, namely approximation power before training, approximation
power after training, and transfer learning capability.

\subsubsection{Approximation Power Before Training}
\label{sec:non-train Fourier}

The first experiment aims to validate the exponential decay of the
approximation error of the \BNetT{} as either the depth $L$ increases
or the number of Chebyshev points $r$ increases.  We construct
and initialize a \BNetT{} to approximate a discrete FT operator,
which has length of input $N = 16,384$ and length of output $K$
representing integer frequency on $[0, K)$. The approximation power
is measured under the relative operator $p$-norm, \ie, $\epsilon_p =
\nicefrac{\norm{\calK - \calB}_p}{\norm{\calK}_p}$, where $\calB$
and $\calK$ denote \BNetT{} and FT operator respectively.

\begin{table}[htp]
\scriptsize
    \centering
    \begin{tabular}{ccccccccc}
        \toprule
        \multirow{2}*{$N$} & \multicolumn{4}{c}{$K=64$}
        & \multicolumn{4}{c}{$K=256$} \\
        \cmidrule(r){2-5}\cmidrule(r){6-9}
        & $L$ & $\epsilon_{1}$ & $\epsilon_{2}$ & $\epsilon_{\infty}$
        & $L$ & $\epsilon_{1}$ & $\epsilon_{2}$ & $\epsilon_{\infty}$ \\
        \midrule
        \multirow{5}*{16384} &
        6  & 3.48\np{e-}02 & 5.25\np{e-}02 & 6.30\np{e-}02 &
        8  & 3.80\np{e-}02 & 7.26\np{e-}02 & 6.94\np{e-}02 \\
        &
        7  & 2.18\np{e-}03 & 4.18\np{e-}03 & 6.36\np{e-}03 &
        9  & 2.39\np{e-}03 & 6.05\np{e-}03 & 6.95\np{e-}03 \\
        &
        8  & 1.37\np{e-}04 & 2.84\np{e-}04 & 5.30\np{e-}04 &
        10 & 1.54\np{e-}04 & 4.31\np{e-}04 & 5.73\np{e-}04 \\
        &
        9  & 8.96\np{e-}06 & 1.79\np{e-}05 & 4.08\np{e-}05 &
        11 & 1.05\np{e-}05 & 2.89\np{e-}05 & 4.37\np{e-}05 \\
        &
        10 & 6.41\np{e-}07 & 1.16\np{e-}06 & 3.11\np{e-}06 &
        12 & 7.64\np{e-}07 & 1.86\np{e-}06 & 3.30\np{e-}06 \\
        \bottomrule
    \end{tabular}
    \caption{Relative error of \BNetT{} before training with $r=4$
    in approximating FT operator.} \label{tab:DFT-result1}
\end{table}

In Table~\ref{tab:DFT-result1}, we fix the number of Chebyshev points
being $r=4$ and varying $L$ for two choices of $K$.  All errors with
respect to different norms decay exponentially as $L$ increases. The
decay rates for different $K$s remain similar, while the prefactor
is slightly larger for larger $K$.

\begin{figure}[ht]
    \scriptsize
    \hspace{-0.5cm}
    \includegraphics[width=2.2in]{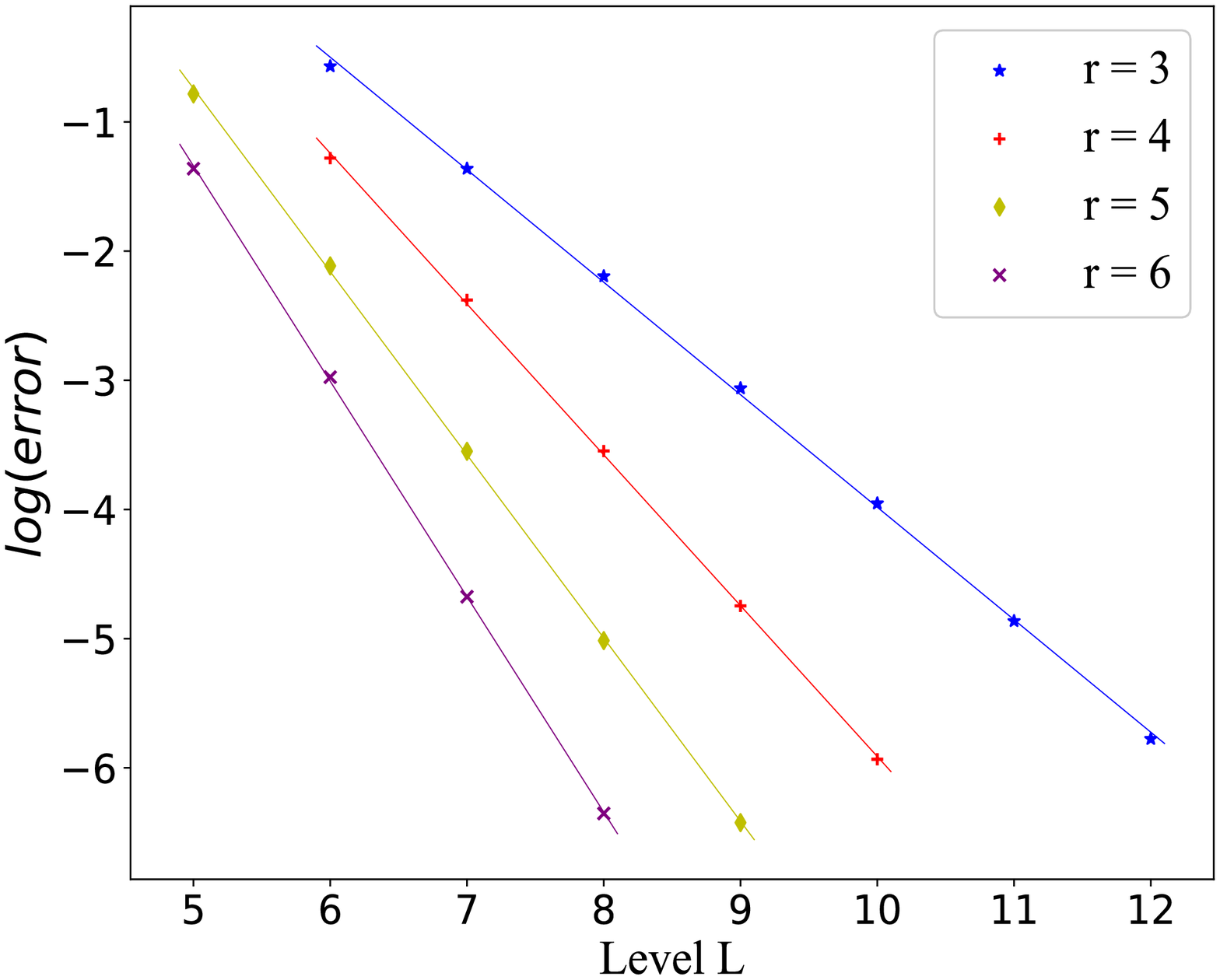} 
    ~
    \raisebox{2.4cm}{
    \begin{tabular}{ccccccccccccc}
        \toprule
        \multirow{2}{*}{$r$} &
        \multicolumn{3}{c}{$K=64$} & \multicolumn{3}{c}{$K=128$} &
        \multicolumn{3}{c}{$K=256$} \\
        \cmidrule(r){2-4} \cmidrule(r){5-7} \cmidrule(r){8-10}
        & $k_1$ & $k_2$ & $k_{\infty}$ &
        $k_1$ & $k_2$ & $k_{\infty}$ & $k_1$ & $k_2$ & $k_{\infty}$ \\
        \toprule
        3 & -0.90 & -0.87 & -0.82
        &-0.90 & -0.85 & -0.82 & -0.91 & -0.85 & -0.82 \\
        4 & -1.18 & -1.14 & -1.04
        &-1.18 & -1.16 & -1.16 & -1.17 & -1.15 & -1.08 \\
        5 & -1.44 & -1.42 & -1.34
        &-1.48 & -1.43 & -1.39 & -1.44 & -1.40 & -1.36 \\
        6 & -1.72 & -1.67 & -1.60
        & -1.72 & -1.69 & -1.60 & -1.72 & -1.70 & -1.61 \\
        \bottomrule
    \end{tabular}
    }
    \caption{
    (Left plot) Exponential convergence rate when $K$ = 64, $p$ = 2.
    (Right table) Convergence rate of \BNetT{} before training in
    approximating FT operator for $r$s. $k_1$, $k_2$, and
    $k_\infty$ are the logarithms of convergence rates under different
    norms, $N=16384$.} \label{tab:DFT-result2}
\end{figure}

In the table in Figure~\ref{tab:DFT-result2}, we calculate the
logarithms of rates of convergence for different $r$s and $K$s
under different norms. The table shows that for all choices of $K$
the convergence rates measured under different norms stay similar
for any fixed $r$. And the convergence rate decreases as $r$ increases.

All of these above convergence behaviors agree with the analysis in
this paper and \cite{Li2019}. And all rates we obtained are better than
the corresponding theoretical ones. In summary, when approximating FT
operator using FT initialized \BNetT{}, the approximation accuracy
decays exponentially as $L$ increases and the rate of convergence
decreases as $r$ increases.

\subsubsection{Approximation Power After Training}
\label{sec:train Fourier}

The second numerical experiment aims to demonstrate the approximation
power of the four networks in approximating FT operator after training.

Each data point used in this section is generated as follows. We first
generate an array of $K$ random complex numbers with each component
being uniformly random in $[-a,a]$. The zero frequency is a random
real number. Second, we apply a Gaussian mask with width (standard
deviation) $G_{\mathrm{width}} = 2$ and center $G_{\mathrm{center}}
=0$(low frequency data) or $G_{\mathrm{center}} =56$(high frequency
data) on the array. The array then is complexly symmetrized to be
a frequency vector and the inverse discrete FT is applied to obtain
the real input vector.  The constant $a$ is chosen such that the two
norm of the output vector is close to $1$. Examples of low and high
frequency input  can be seen in Appendix~\ref{sec:EMR}.

In this experiment, we have input length being $N = 128$, output
length being $K = 8$, level number being $L = 5$, channel parameter
being $r=3$.  All networks are trained under the infinity data setting,
\ie, the training data is randomly generated on the fly. ADAM optimizer
with batch size $256$ and exponential decay learning rate is adopted.
For FT initialized networks, the maximum training steps is 10,000,
whereas for random initialization we train 20,000 steps.  The reported
relative error in vector two norm is calculated on a testing data
set of size $16,384$ with the same distribution as the training data
set. Default values are used for any unspecified hyper parameters.

\begin{table}[ht]
\scriptsize
    \centering
    \begin{tabular}{lccccc}
        \toprule
        \multirow{2}{*}{Network}
        & \multirow{2}{*}{$\sharp$ Parameters}
        & \multicolumn{2}{c}{Low Frequency}
        & \multicolumn{2}{c}{High Frequency}\\
        && Pre-Train Rel Err & Test Rel Err 
        & Pre-Train Rel Err & Test Rel Err\\
        \toprule
        \BNetT{}-FT & \multirow{2}{*}{9252}
        &   2.44\np{e-}3 & 1.33\np{e-}5
        &   2.61\np{e-}3 & 1.29\np{e-}5\\
        \BNetT{}-rand &
        & 1.38\np{e+}0 & 8.82\np{e-}3 
        & 1.25\np{e+}0 & 8.50\np{e-}3\\
        \midrule
        CNN-FT & \multirow{3}{*}{49572}
        & 2.44\np{e-}3 & 9.29\np{e-}6
        & 2.61\np{e-}3 & 7.54\np{e-}6 \\
        CNN-rand &
        & 5.09\np{e+}0 & 4.63\np{e-}2 
        & 2.73\np{e+}0 & 2.20\np{e-}2\\
        CNN-BNet2(FT-trained) &
        & 1.33\np{e-}5 & 6.18\np{e-}6 
        & 1.29\np{e-}5 & 4.06\np{e-}6\\
        \bottomrule
    \end{tabular}
    \caption{Relative errors of networks in approximating FT operator
    before and after training. The last row use the trained parameters
    in the first row as it's initialization.} \label{tab:approx-bnet}
\end{table}

Table~\ref{tab:approx-bnet} shows the pre-training and testing
relative errors for \BNetT{}-FT, \BNetT{}-rand, CNN-FT, CNN-rand and
CNN-\BNetT{}(FT-trained) on both low and high frequency training set.
Comparing the results, every network have similar performance on
both data set, \BNetT{} and CNN have similar performance for both
initializations, while \BNetT{} has only about $\nicefrac{1}{5}$
parameters as CNN. Hence those extra coefficients in CNN do not
improve the approximation to FT operator. On the other hand, FT
initialization achieves an accuracy better than that of \BNetT{}-rand
and CNN-rand after training. After training FT initialized networks,
extra two digits accuracy is achieved for both \BNetT{}-FT and
CNN-FT. The CNN with trained FT initialized \BNetT{} performs slightly
better than CNN-FT, but the improvement is not as significant as FT
initialization. We conjecture that the local minima found through the
training from the FT initialization has a narrow and deep well on the
energy landscape such that the random initialization with stochastic
gradient descent is not able to find it efficiently.

\subsubsection{Transfer Learning Capability}
\label{sec:transfer}

This numerical experiment compares the transfer learning capability
of four networks. The training and testing data are generated in
a same way as in Section~\ref{sec:train Fourier} with different
choices of $G_{\mathrm{center}}$ and $G_{\mathrm{width}}$. We have
three training sets: low frequency training set ($G_{\mathrm{center}}
= 0$ and $G_{\mathrm{width}} = 2$), high frequency training set
($G_{\mathrm{center}} = 7$ and $G_{\mathrm{width}} = 2$) and mixture
training set (no Gaussian mask).  A sequence of testing sets of size
$16,384$ are generated with $G_{\mathrm{center}} = 0, 0.2, 0.4, \dots,
7$ and $G_{\mathrm{width}} = 2$.

The networks used here have the same structure and hyper-parameters
as in Sec~\ref{sec:train Fourier} while the channel parameter $r =
2$ instead of $3$ here.  Each experiment is repeated $20$ times. Then
the mean and standard deviation of the error in two norm are reported
below.

\begin{figure}[ht]
    \centering \subfigure[low frequency training set]{
        \includegraphics[width=1.9in]
        {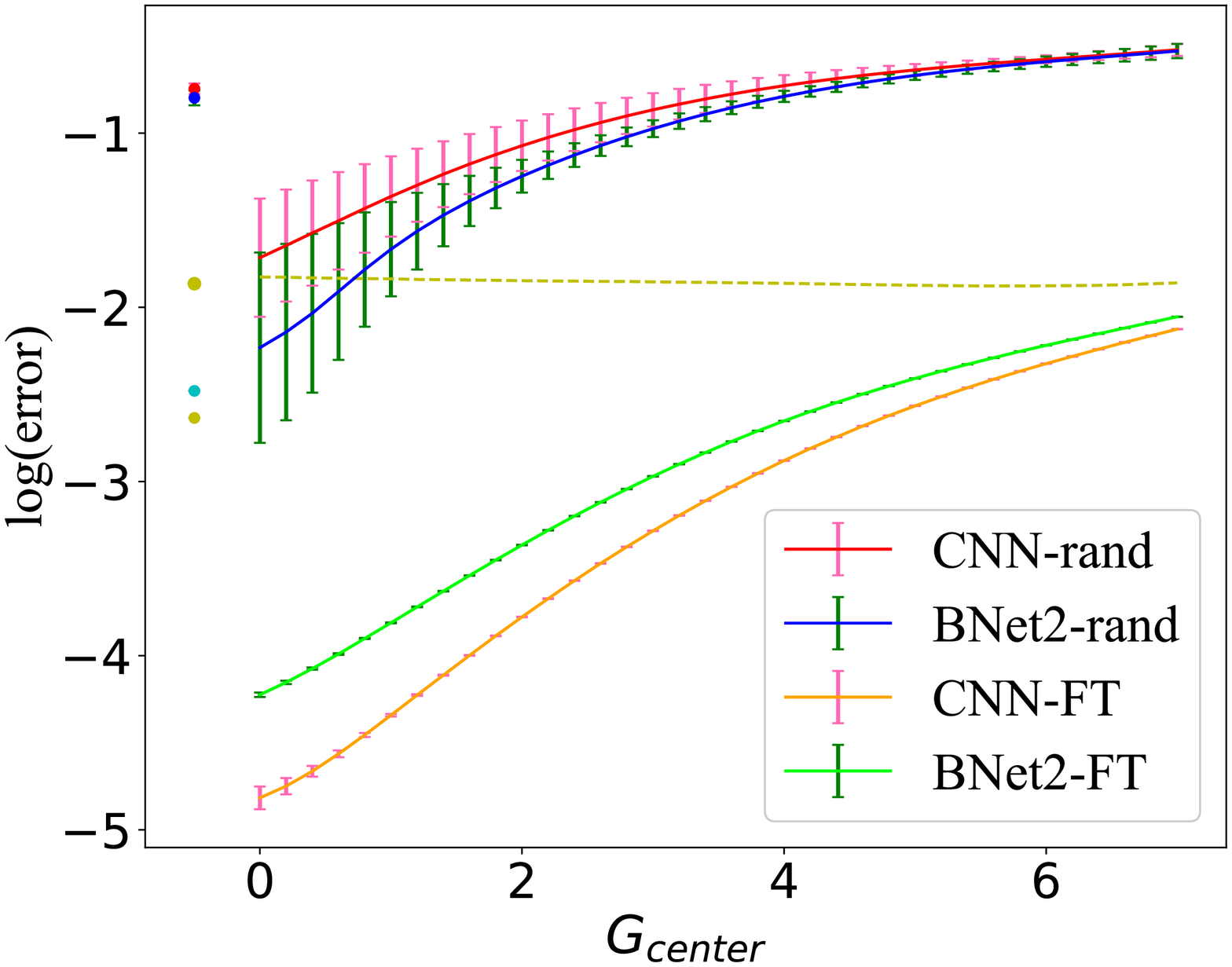}}
     \subfigure[high frequency training set]{
        \includegraphics[width=1.9in]
        {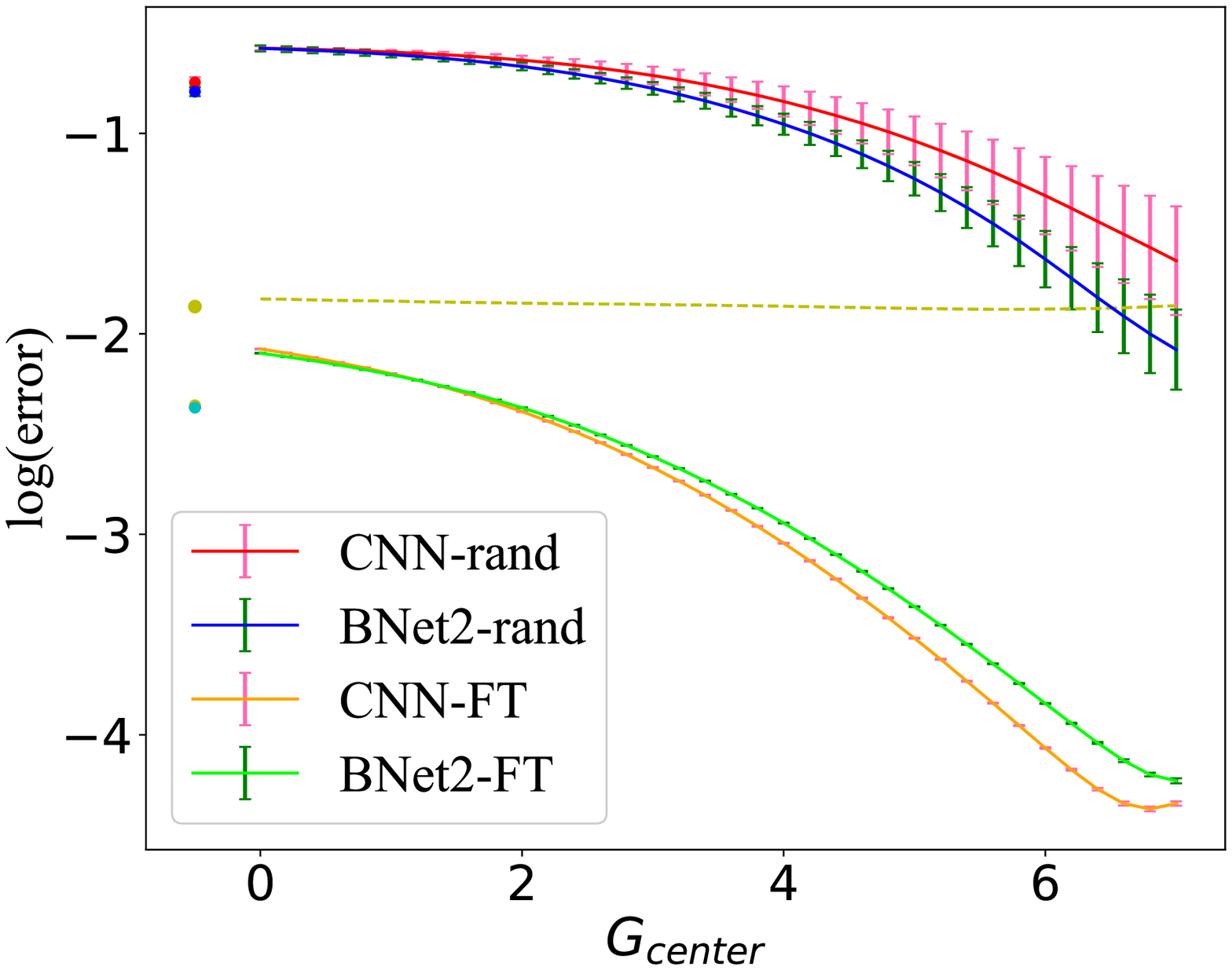}}
     \subfigure[mixture training set]{
        \includegraphics[width=1.92in]
        {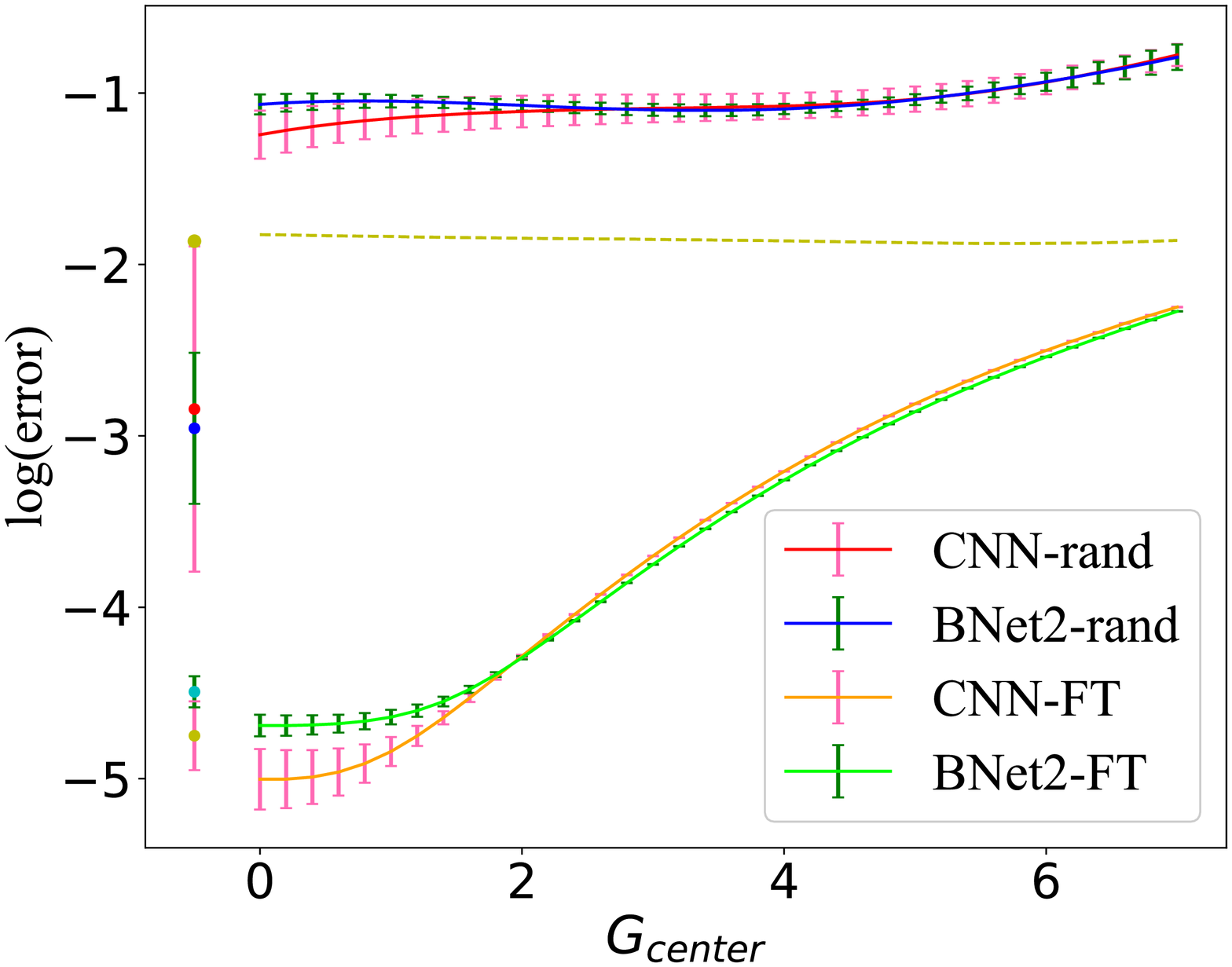}}
    \caption{Figures show the transfer learning results of four
    networks trained on three different training sets.  The horizontal
    axis represents testing sets with different $G_{\mathrm{center}}$.
    The testing results on the mixture testing set are plotted as
    the isolated error bars at the left of each plot.  Each error bar
    represents the mean and standard deviation across $20$ repeating
    experiments. The horizontal dash lines indicate the testing error
    of FT initialized networks before training.} \label{fig:translearn}
\end{figure}

As in Figure~\ref{fig:translearn}, for both initializations, \BNetT{}
and CNN have similar accuracy especially on testing sets away from
the training set. Taking the FT initialization before training
as a reference, we also notice that even if randomly initialized
networks can reach the accuracy of the reference on some testing sets,
they lose accuracy on transferred testing sets. On the other side,
FT initialized networks after training maintain accuracy better
than that of the reference on all testing sets.  In terms of the
stability after training, \BNetT{}-FT and CNN-FT are much more stable
than \BNetT{}-rand and CNN-rand, which is due to the randomness in
initializers. This phenomenon also emphasizes the advantage of FT
initialization in stability and repeatability.

\subsection{Energy and Solution Map of Elliptic PDEs}
\label{subsec:exp-PDEs}

This section focus on the elliptic PDE of the following form,
\begin{equation} \label{eq:ellipticPDE}
    -\frac{\diff}{\diff x}\Big( a(x) \frac{\diff u(x)}{\diff x} \Big)
    + b u^3(x) = f(x), \quad x \in [0,1),
\end{equation}
with periodic boundary condition, where $a(x) > 0$ denotes coefficients
and $b$ denotes the strength of nonlinearity. Such an equation appears
in a wide range of physical models governed by Laplace's equation,
Stokes equation, etc.  Equation \eqref{eq:ellipticPDE} is discretized
on a uniform grid with $N$ points.

\subsubsection{Energy of Laplace Operator}
\label{sec:energyLaplace}

In this section, we aim to construct an approximation of the energy
functional of 1D Poisson's equations, \ie, $a(x) \equiv 1$ and $b=0$.
The energy functional of Poisson's equation is defined as the negative
inner product of $u$ and $f$, which can also be approximated by
a quadratic form of the leading low-frequency Fourier components.
Hence, Here we adopt \BNetT{}-FT, \BNetT{}-rand, CNN-FT, and CNN-rand
with an extra square layer, which is called task-dependent layer.

In this numerical example, the input $f$ has the same distribution as
that in Section~\ref{sec:train Fourier}. All other hyper parameters
of networks and the training setting are also identical to that in
Section~\ref{sec:train Fourier}.

\begin{table}[htp]
\scriptsize
    \centering
    \begin{tabular}{lccc}
        \toprule
        &$\sharp$ Parameters &
        Pre-Train Rel Err & Test Rel Err \\
        \midrule
        \BNetT{}-FT&\multirow{2}{*}{9268}
        & 2.11\np{e-}3 & 8.10\np{e-}6 \\
        \BNetT{}-rand& & 7.97\np{e-}1 & 4.62\np{e-}3 \\
        \cmidrule(r){1-4}
        CNN-FT&\multirow{2}{*}{49588}
        & 2.11\np{e-}3 & 4.79\np{e-}6 \\
        CNN-rand&  & 5.53\np{e-}1 & 6.21\np{e-}3 \\
        \bottomrule
    \end{tabular}
    \caption{Training results for networks in representing the energy
    of the Laplace operator.} \label{tab:approx-energy}
\end{table}

Table~\ref{tab:approx-energy} shows the results for energy
of 1D Laplace operators, which has similar property as
Table~\ref{tab:approx-bnet}.  Hence all conclusions in
Section~\ref{sec:train Fourier} apply here.

\subsubsection{End-to-end Linear Elliptic PDE Solver}
\label{sec:linearequation}

In this section, we aim to represent the end-to-end solution map
of linear elliptic PDEs by an encoder-decoder structure. The linear
elliptic PDE is \eqref{eq:ellipticPDE} with high contrast coefficient
$a(x)$ as,
\begin{equation} \label{eq:a}
    a(x_i) =
    \begin{cases}
        10, & \floor{ \frac{8i-N}{2N} } \equiv 0\, (\mathrm{mod} \, 2) \\
        1, & \floor{ \frac{8i-N}{2N}  } \equiv 1\, (\mathrm{mod} \, 2) \\
    \end{cases},
\end{equation}
for $x_i$ being the uniform point in $[0,1)$ and $b = 0$.

It is well known that the inverse of linear constant coefficient
Laplace operator can by represented by $\calF^\star \calD^{-1}
\calF$ where $\calF$ denotes the FT and $\calD$ is a diagonal
operator. Therefore, we design our network in the same sprite. Our
network contains three parts: a \BNetT{}/CNN encoder with input
length $2N$,output length $K_{en}$, a $K_{en} \times K_{de}$ fully
connected dense layer with bias terms and activation function, and a
\BNetT{}/CNN decoder with input length $K_{de}$, output length $N$.
Since the input $f$ is real function, here we apply odd symmetry to
it and initialize the first \BNetT{}/CNN as FT to make the first
part of the network serves as sine transform. Then we initialize $D$
according to $a(x)$, and initialize the third part to be an inverse
sine transform so that the overall network is an approximation of
the inverse operator.  In this example, we set $N = 64$, $K_{en} =
8$, and $K_{de} = 16$. Both \BNetT{}s/CNNs are constructed with $L=4$
layers and channel parameter $r=3$.

Each training and testing data is generated as follows.  We first
generate an array of length $K$ with $K-1$ random numbers. The first
entry is fixed to be 0 to incorporate the periodic boundary condition,
whereas the following $K-1$ entries are uniform sampled from $[-1,1]$.
Then an inverse discrete sine transform is applied to obtain the input
vector. The reference solution is calculated through traditional
spectral methods on a finer grid of $16N$ nodes. The training and
testing data set contain $4,096$ and $5,000$ points respectively.
Other settings are the same as in Section~\ref{sec:train Fourier}.

\begin{table}[ht]
    \scriptsize
    \centering
    \begin{tabular}{lccccccc}
        \toprule
        & \multirow{2}{*}{$\sharp$ Parameters} &
        \multicolumn{3}{c}{Linear PDE} &
        \multicolumn{3}{c}{Nonlinear PDE} \\
        &
        & Pre-Train Rel Err &Train Rel Err & Test Rel Err
        & Pre-Train Rel Err &Train Rel Err & Test Rel Err \\
        \toprule
        \BNetT{}-FT & \multirow{2}{*}{17856}
        & 5.16\np{e-}2 & 4.71\np{e-}3 & 4.86\np{e-}3
        & 3.48\np{e+}0 & 1.97\np{e-}2 & 2.02\np{e-}2\\
        \BNetT{}-rand &
        & 9.75\np{e+}0 & 4.26\np{e-}2 & 4.43\np{e-}2
        & 4.37\np{e+}2 & 1.00\np{e+}0 & 1.00\np{e+}0\\
        \midrule
        CNN-FT & \multirow{2}{*}{82368}
        & 5.16\np{e-}2 & 3.80\np{e-}3 & 3.96\np{e-}3
        & 3.48\np{e+}0 & 1.36\np{e-}2 & 1.52\np{e-}2\\
        CNN-rand &
        & 3.53\np{e+}0 & 2.02\np{e-}2 & 2.03\np{e-}2
        & 5.65\np{e+}2 & 1.00\np{e+}0 & 1.00\np{e+}0 \\
        \bottomrule
    \end{tabular}
    \caption{Relative errors in approximating the solution map of
    the linear and nonlinear elliptic PDE.} \label{tab:approx-lede}
\end{table}

\begin{figure}[ht]
    \centering
    \includegraphics[height=0.3\textwidth,width=0.8\textwidth]{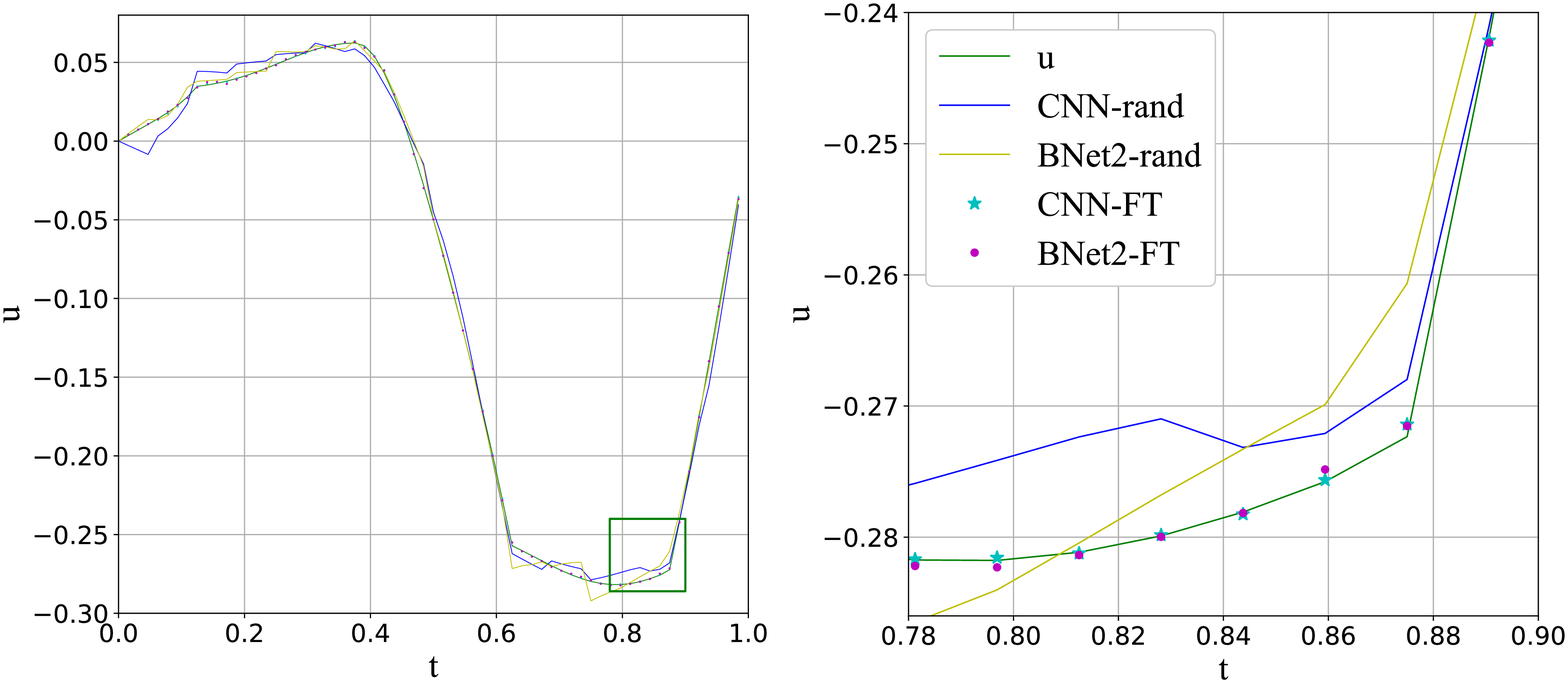}
    \caption{The left figure shows an example solution $u$ and the
    output from four networks for the linear elliptic PDE. The
    right figure is a zoom-in of the green box in the left.}
    \label{fig:linear_ode}
\end{figure}

Table~\ref{tab:approx-lede} and Figures~\ref{fig:linear_ode} show
that in this end-to-end task, CNN performs slightly better than
\BNetT{} at the cost of $5$ times more parameters. Training from FT
initialization in both cases provides one more digit of accuracy.
Figures~\ref{fig:linear_ode} further shows that \BNetT{}-FT and CNN-FT
significantly outperform \BNetT{}-rand and CNN-rand near sharp changing
areas in $u$.

\subsubsection{End-to-end Nonlinear Elliptic PDE Solver}
\label{sec:nonlinearequation}

In this section, we focus on a highly nonlinear elliptic PDE
\eqref{eq:ellipticPDE} with $a(x) \equiv 1$ and $b = 10^3$.
The reference solution for nonlinear PDEs in general is difficult and
expensive to obtain. Hence, in this section, we apply the solve-train
framework proposed in~\cite{li2019variational} to avoid explicitly
solving the nonlinear PDE.

Denoting the nonlinear PDE as an operator acting on $u$, \ie, $\calA(u)
= f$, our loss function here is defined as
\begin{equation} \label{eq:loss}
    \ell \Big( \{f_i\}_{i=1}^{\Ntrain}, \calA, \calN \Big)
    = \frac{1}{\Ntrain} \sum_{i=1}^{\Ntrain} \norm{f_i - \calA
    \big( \calN (f_i) \big)}^2,
\end{equation}
where $\calN$ denotes the used neural network. The reported relative
error is calculated on testing data $\{g_i\}_{i=1}^{\Ntest}$
as follows,
\begin{equation} \label{eq:relerr}
    \frac{1}{\Ntest} \sum_{i=1}^{\Ntest} \frac{\norm{g_i - \calA
    \Big( \calN(g_i) \Big)}}{\norm{g_i}}.
\end{equation}

The same networks and other related settings as in
Section~\ref{sec:linearequation} are used here.

Table~\ref{tab:approx-lede} shows that under solve-train framework
randomly initialized networks are not able to converge to a meaningful
result, whereas FT initialized networks find a representation for
the solution map with $2$ digits accuracy. Partially, this is due
to the extra condition number of $\calA$ introduced by solve-train
framework in training. Comparing \BNetT{}-FT with CNN-FT, we find
similar conclusions as before, \ie, CNN-FT achieves slightly better
accuracy with higher cost in the number of parameters.

\subsection{Denoising and Deblurring of 1D Signals}
\label{subsec:exp-denoise-deblur}

In this section, we aim to apply networks to the denoising and
deblurring tasks in signal processing. An encoder-decoder structure
is used in this experiment, which concatenates two networks, \ie,
two \BNetT{}-FT, two \BNetT{}-rand, two CNN-FT or two CNN-rand. Such
a structure with FT initialization reproduces a low pass filter.

The low frequency true signal $f$ is generated as the input
vector in Section~\ref{sec:train Fourier}. Two polluted signals,
$f_{\mathrm{noise}}$ and $f_{\mathrm{blur}}$, are generated by adding
a Gaussian noise with standard deviation $0.002$ and convolving a
Gaussian with standard deviation $3$, respectively.  The mean relative
errors of $f_{\mathrm{noise}}$ and $f_{\mathrm{blur}}$ are $0.0226$
and $0.165$ respectively.

Regarding the encoder-decoder structure, the first part has input
length $N = 128$ and output length $K = 8$ in representing frequency
domain $[0,K)$.  After that, the output of the first part is complex
symmetrized to frequency domain $[-K,K)$.  Then the second part
has input length $16$, output length $128$. In both parts, we adopt
$L=4$ layers.  The other hyper parameters are the same as that in
Section~\ref{sec:train Fourier}.  All relative errors are measured
in two norm.

\begin{table}[ht]
    \scriptsize
    \centering
    \begin{tabular}{lccccc}
        \toprule
        \multirow{2}{*}{Network} & \multirow{2}{*}{$\sharp$ para} &
        \multicolumn{2}{c}{$f_{\mathrm{noise}}$} &
        \multicolumn{2}{c}{$f_{\mathrm{blur}}$} \\
        \cmidrule(r){3-4}
        \cmidrule(r){5-6}
        & & Pre-Train Rel Err & Test Rel Err
        & Pre-Train Rel Err & Test Rel Err \\
        \midrule
        \BNetT{}-FT   & \multirow{2}{*}{19,392} &
        9.56\np{e-}2 & 7.54\np{e-}3 & 1.64\np{e-}1 & 8.02\np{e-}4 \\
        \BNetT{}-rand & &
        1.07\np{e+}0 & 1.52\np{e-}2 & 1.02\np{e+}0 & 1.07\np{e-}2 \\
        \midrule
        CNN-FT & \multirow{2}{*}{83,904} &
        9.56\np{e-}2 & 7.74\np{e-}3 & 1.64\np{e-}1 & 8.19\np{e-}4 \\
        CNN-rand & &
        1.23\np{e+}0 & 1.28\np{e-}2 & 1.05\np{e+}0 & 9.95\np{e-}3 \\
        \bottomrule
    \end{tabular}
    \caption{ Relative error of denoising and deblurring of 1D signals.}
    \label{tab:approx-denobl}
\end{table}

\begin{figure}[ht]
    \centering
    \subfigure[Example of signal denoising]{
    \includegraphics[width=0.8\textwidth]{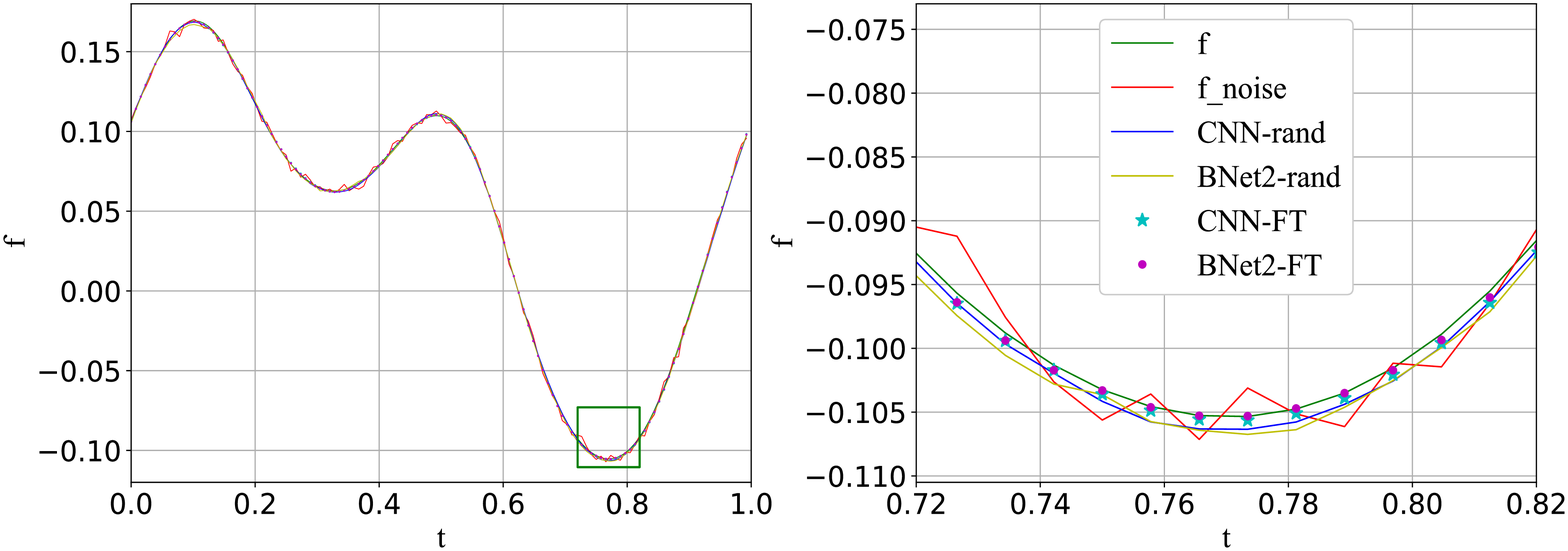}}
    \subfigure[Example of signal deblurring]{
    \includegraphics[width=0.8\textwidth]{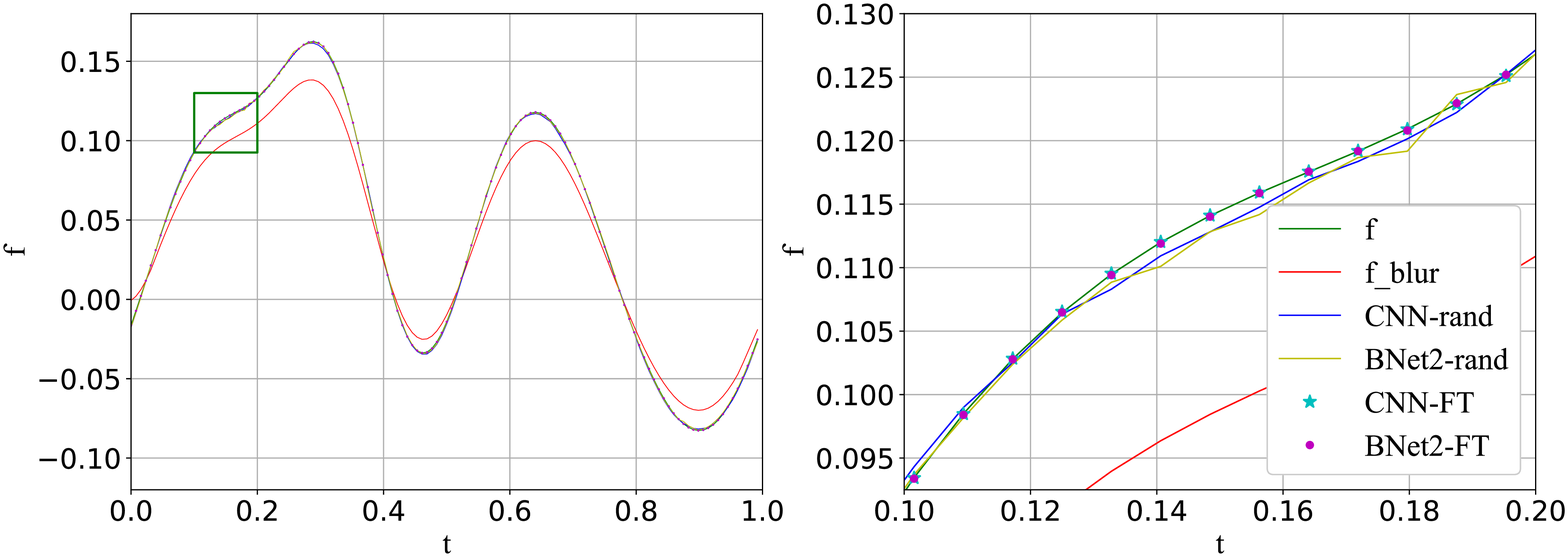}}
    \caption{(a) and (b) show an example of denoising and deblurring
    respectively. The right figures are zoom-in of green boxes in the
    left figures.} \label{fig:denoise/deblur}
\end{figure}

Table~\ref{tab:approx-denobl} lists all relative errors of four
networks and Figure~\ref{fig:denoise/deblur} shows the performance of
four networks on an example signal. We observe that, for both tasks,
the FT initialized networks have better accuracy than their randomly
initialized counterparts. Under the same initialization, \BNetT{}
achieves similar accuracy as CNN with much fewer parameters. Comparing
two tasks, we notice that the improvement of FT initialization over
random initialization is more significant on deblurring task than
that on denoising task. For denoising tasks, as we enlarge additive
noise level, \BNetT{}-rand and CNN-rand perform almost as good as
\BNetT{}-FT and CNN-FT. However, for deblurring tasks, we always
observe significant improvement from FT initialization.

\section*{Acknowledgement}

The authors thank Bin Dong,  Yiping Lu for discussion on image inverse
problems. The work of YL is supported in part by National Science
Foundation via grants DMS-1454939 and ACI-1450280.  XC is partially
supported by NSF (DMS-1818945, DMS-1820827), NIH (grant R01GM131642)
and the Alfred P. Sloan Foundation.

\bibliographystyle{abbrvnat}
\bibliography{paper}

\appendix

\section{BNet Revisit}\label{sec:BNet}

For a $L$ layer BNet with an additional parameter $L_t$ denoting
the number of layers before switch layer, the feedforward network is
as follows.
\begin{itemize}
    \item \textbf{Layer 0: } The same as Layer 0 in CNN.

    \item \textbf{Layer $\ell = 1, \dots, L_t-1$: } The same as Layer
    $\ell$ in \BNetT{}.
    
    \item \textbf{Switch Layer:} This layer first applies many small
    dense layer to each part and then the role of spacial dimension and
    channel dimension is switched afterwards.  We denote the hidden
    variables on switch layer as $Z^{(s)}$. The connection between
    the $\ell$-th layer and the $(s)$-th layer hidden variables obeys,
    \begin{equation}
        \Z{s}{i}{rj+c} = \sigma \Big(\B{s}{i,j,c} + \sum_{k \in
        [r]}
        \D{s}{i}{j}{c}{k} \Z{L_t}{i}{rj+k} \Big),
    \end{equation}
    for $j \in [2^{L_t}]$, $i \in [\frac{N}{2^{L_t} w}]$, and $c \in
    [r]$.  Here $D^{(s)}$ and $B^{(s)}$ denote the weights and bias
    respectively.
    
    \item \textbf{Layer $\ell = L_t, \dots, L-1$: } The $2^\ell
    r$ in-channels are equally partitioned into $2^\ell$ parts.
    A 1D transposed convolutional layer is applied.  The connection
    between the $\ell$-th layer and the $(\ell+1)$-th layer hidden
    variables obeys,
    \begin{equation}
        \Z{\ell+1}{2j+i}{c} = \sigma \Big(\B{\ell}{c} + \sum_{k
        \in [2^{L-\ell} r]^{2^{L-\ell-1}}_p} \W{\ell}{i}{k}{c}
        \Z{\ell}{j}{k} \Big),
    \end{equation}
    for $j \in [\frac{N}{2^{L - \ell} w}]$, $i \in [2]$,
    $c \in [2^{L-\ell-1} r]^{2^{L-\ell-1}}_p$, and $p \in
    [2^{L-\ell-1}]$. Here we
	abuse notation $Z^{(L/2)} = Z^{(s)}$.

    \item \textbf{Layer $L$: } The last layer links the $L$-th layer
    hidden variables with the output $Y$, \ie,
    \begin{equation} 
        \Y{c} = \sum_{k \in
        [r]} \sum_{i \in [\frac{N}{w}]^{2^L}_p}
        \W{L}{i}{k}{c} \Z{L}{i}{k},
    \end{equation}
	for $c \in [K]^{2^L}_p$ and $p \in [2^L]$.
    
\end{itemize}

\section{Complex valued network}
\label{sec:complexnetwork}

In order to realize complex number multiplication and addition via
nonlinear neural network, we first represent a complex number as four
real numbers, i.e., a complex number $x = \Re x+\imath \Im x \in \bbC$
is represented as
\begin{equation}
    \begin{pmatrix}
        (\Re x)_+ & (\Im x)_+ & (\Re x)_- & (\Im x)_-
    \end{pmatrix}^\top,
\end{equation}
where $(z)_+ = \max(z,0)$ and $(z)_- = -\min(z,0)$ for any $z \in
\bbR$. The vector form of $x$ contains at most two nonzeros. The
complex number addition is the vector addition directly, while
the complex number multiplication must be handled carefully. Let
$a,x \in \bbC$ be two complex numbers. The multiplication $y = ax$
is produced as the activation function acting on a matrix vector
multiplication, \ie,
\begin{equation} \label{eq:exeq}
    \sigma \left(
    \begin{pmatrix}
        \Re a & -\Im a & -\Re a & \Im a \\
        \Im a & \Re a & -\Im a & -\Re a \\
        -\Re a & \Im a & \Re a & -\Im a \\
        -\Im a & -\Re a & \Im a & \Re a \\
    \end{pmatrix}
    \begin{pmatrix}
        (\Re x)_+ \\
        (\Im x)_+ \\
        (\Re x)_- \\
        (\Im x)_-
    \end{pmatrix}
    \right)
    =
    \begin{pmatrix}
        (\Re y)_+ \\
        (\Im y)_+ \\
        (\Re y)_- \\
        (\Im y)_-
    \end{pmatrix}.
\end{equation}
In the initialization, all prefixed weights are in the role of $a$
instead of $x$. In order to simplify the description below, we define
an extensive assign operator as $\exeq$ such that the 4 by 4 matrix $A$
in \eqref{eq:exeq} then obeys $A \exeq a$. Without loss of generality,
\eqref{eq:exeq} can be extended to complex matrix-vector product and
$\exeq$ notation is adapted accordingly as well.

\section{Sketch Proof of Theorem~\ref{thm:matrix-approx}}
\label{sec:proof-matrix-approx}

The detail proof of Theorem~\ref{thm:matrix-approx} is composed of
layer by layer estimations on the multiplicative weight matrices,
which is analogy to the proof of Theorem~4.8 in~\cite{Li2019}.
Here we omit the detail and discuss the relations and differences.

If we consider the approximation under condition $L \leq \log K$, then
the bound in Theorem~\ref{thm:matrix-approx} is exactly the same as
that in Theorem~4.8 in~\cite{Li2019} with $L_\xi = 0$, where $L_\xi$
denotes the number of Conv-T layers after switch layer. However,
when we consider $L > \log K$, the number of partitions of the
frequency domain in BNet is limited by $K$ due to the existence of
switch layer. Hence the bottleneck domain pair, $A$ and $B$ as in
Proposition~\ref{thm:low-rank}, are of length $1$ and $1$ respectively.
The Chebyshev interpolation error is then
\begin{equation}
    \left(2 + \frac{2}{\pi} \ln r\right)\left( \frac{\pi e}{2r}
    \right)^r,
\end{equation}
which can be well controlled as we increase $r$. Therefore, in
Theorem~4.8 in~\cite{Li2019}, the approximation error is also
controlled in terms of $r$.

\BNetT{}, different from BNet, dose not have the constraint from
switch layer. The frequency domain can be partitioned into $2^L$
subdomains and each has length $\frac{K}{2^L}$. When the number of
subdomains is larger than the number of output frequencies, only
those subdomains containing output frequencies are constructed in the
network and considered in the proof. Due to the fine partition of the
frequency domain, in Proposition~\ref{thm:low-rank}, the product of the
lengths of domain pair $A$ and $B$ is always bounded by $\frac{K}{2^L}$
and the Chebyshev interpolation error is
\begin{equation} \label{eq:cheberror}
    \left(2 + \frac{2}{\pi} \ln r\right)\left( \frac{\pi e K}{r 2^{L+1}}
    \right)^r
\end{equation}
for any $L$. Replacing the interpolation error in the proof of
Theorem~4.8 in \cite{Li2019} by \eqref{eq:cheberror} throughout layers
proves Theorem~\ref{thm:matrix-approx}.

\section{BNet against BNet2}\label{sec:BNet1vs2}

This numerical experiment compares \BNetT{} with BNet in a similar
task as in Section~\ref{sec:train Fourier}. Comparing to that in
Section~\ref{sec:train Fourier}, $L = 3$ are used here, while $N$,
$K$ and $r$ remain the same. We train both BNet and \BNetT{} on the
same training set with the same training hyper-parameters.

\begin{table}[ht]
\scriptsize
    \centering
    \begin{tabular}{lccccc}
        \toprule
        &$\sharp$ Parameters & training time&
        Pre-Train Rel Err & testing time& Test Rel Err \\
        \toprule
        \BNetT{}-FT&\multirow{2}{*}{4596}
        &\multirow{2}{*}{68.3 s}& 1.28\np{e-}1 
        &\multirow{2}{*}{0.095 s}& 3.13\np{e-}4 \\
        \BNetT{}-rand& & &1.06\np{e+}0 & &1.69\np{e-}2 \\
        \cmidrule(r){1-6}
        BNet-FT&\multirow{2}{*}{3876}
         &\multirow{2}{*}{95.4 s}& 8.01\np{e-}2 
         &\multirow{2}{*}{0.182 s}& 3.51\np{e-}4 \\
        BNet-rand& & & 1.03\np{e+}0 & &1.84\np{e-}2 \\
        \bottomrule
    \end{tabular}
    \caption{Training results for BNet and \BNetT{}} \label{tab:Bnet1vs2}
\end{table}

Due to the huge difference in architectures of BNet and \BNetT{},
the numbers of parameters and pre-train relative errors are not
identical but stay close to each other. After training, \BNetT{}
achieves slightly better accuracy than that of BNet for both random
initialization and FT initialization. This is likely due to the
small difference in the number of parameters. Regarding the runtime,
both the training and evaluation of BNet are more expensive than that
of \BNetT{}.

\section{Extra Numerical Results}
\label{sec:EMR}

\begin{figure}[H]
    \centering \subfigure[Low frequency input example]{
    \includegraphics[width=0.48\textwidth]{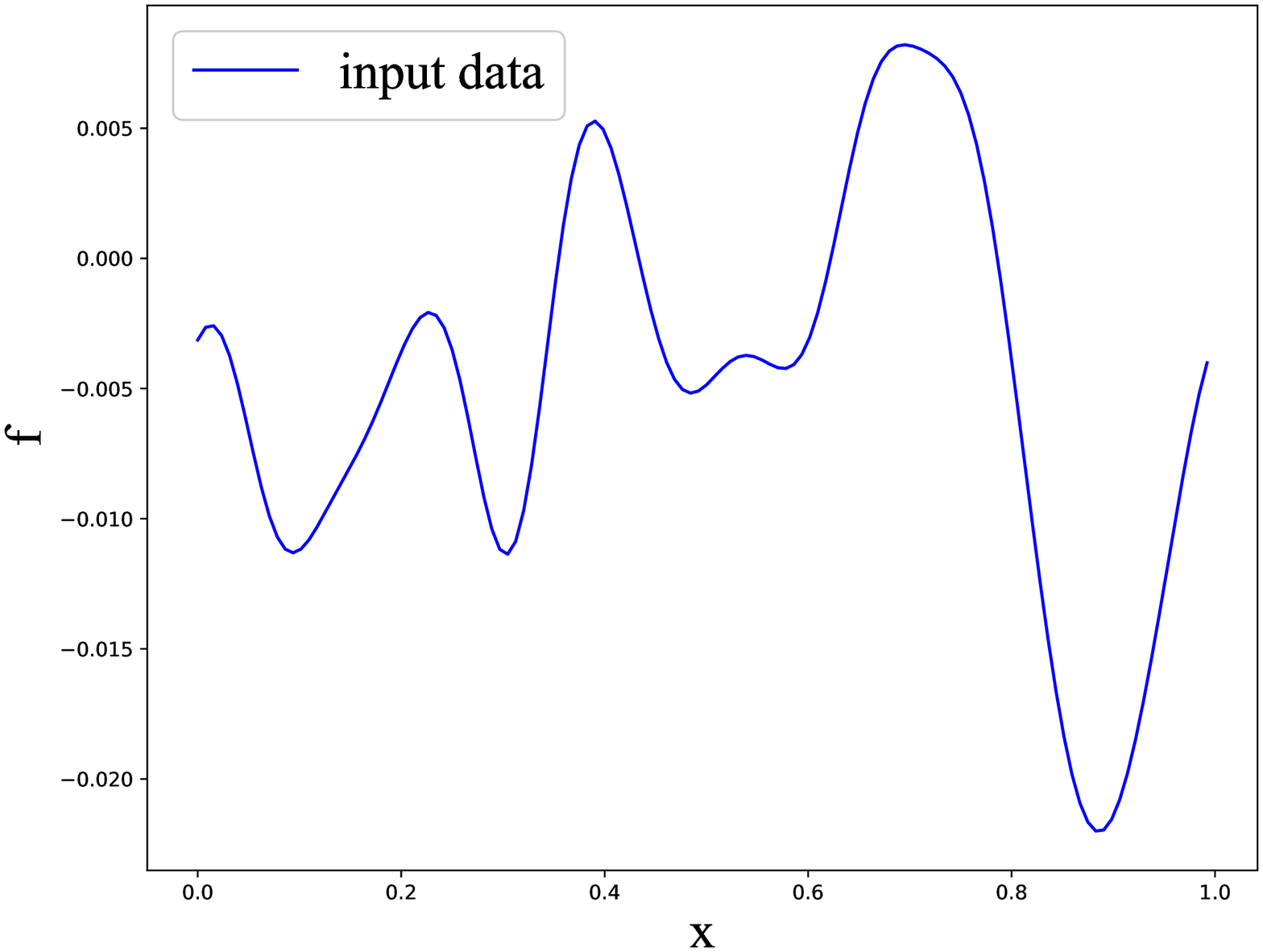}}
    \subfigure[High frequency input example]{
    \includegraphics[width=0.48\textwidth]{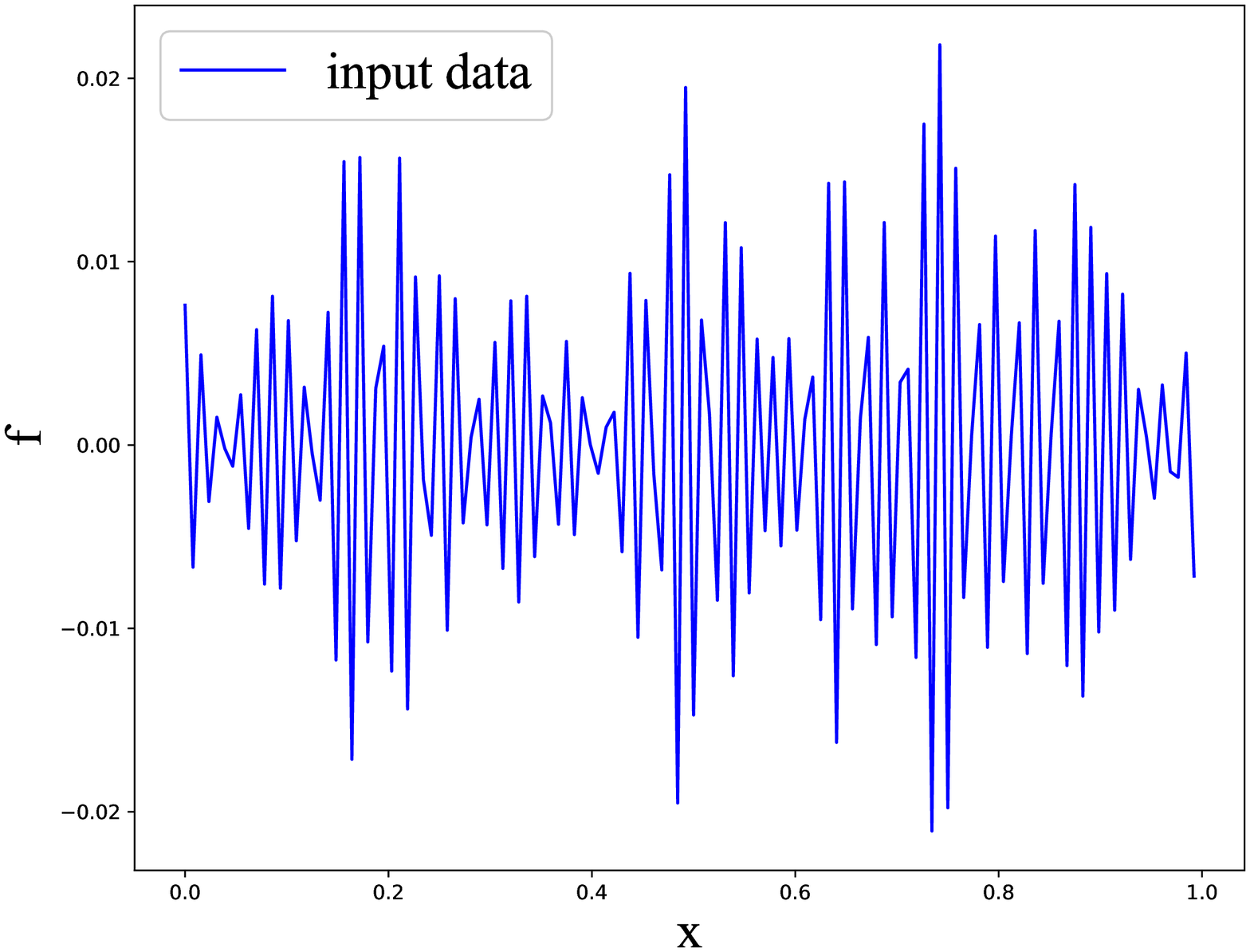}}
    \caption{(a) and (b) show examples of low and high frequency
    input used in Section~\ref{sec:train Fourier}.} \label{fig:lh}
\end{figure}

\end{document}